%% file: farinati.tex
\documentclass[runningheads]{llncs}

\usepackage{graphicx}%
\usepackage{multirow}%
\usepackage{amsmath,amssymb,amsfonts}%
\usepackage{mathrsfs}%
\usepackage[title]{appendix}%
\usepackage{xcolor}%
\usepackage{textcomp}%
\usepackage{manyfoot}%
\usepackage{booktabs}%
\usepackage{algorithm}%
\usepackage{algorithmicx}%
\usepackage{algpseudocode}%
\usepackage{listings}%
\usepackage{glossaries}
\usepackage{hyperref}
\usepackage{cleveref}
\usepackage{subcaption}
\usepackage{cancel}
\usepackage[inline]{enumitem}

\usepackage[utf8]{inputenc}

\usepackage{color,colortbl,etoolbox}
\usepackage[linewidth=1pt,ticklabelscale=0.75,titleyshift=-1mm]{plotmacros}

\usetikzlibrary{external}

\usepackage{caption}
\let\llncssubparagraph\subparagraph
\let\subparagraph\paragraph 
\usepackage[compact]{titlesec}         
\let\subparagraph\llncssubparagraph    
\titlespacing{\section}{0pt}{1pt}{1pt} 
\titlespacing{\subsection}{0pt}{1pt}{1pt}
\titlespacing{\subsubsection}{0pt}{1pt}{1pt}
\titlespacing{\paragraph}{0pt}{1pt}{1pt}

\usepackage{comment}

\newacronym{ml}{ML}{Machine Learning}
\newacronym{ec}{EC}{Evolutionary Computation}
\newacronym{gp}{GP}{Genetic Programming}
\newacronym{gsgp}{GSGP}{Geometric Semantic Genetic Programming}
\newacronym{gso}{GSO}{Geometric Semantic Operator}
\newacronym{gsm}{GSM}{Geometric Semantic Mutation}
\newacronym{igsm}{IGSM}{Inflate Geometric Semantic Mutation}
\newacronym{dgsm}{DGSM}{Deflate Geometric Semantic Mutation}
\newacronym{gsc}{GSC}{Geometric Semantic Crossover}
\newacronym{slim}{SLIM-GSGP}{Semantic Learning algorithm based on Inflate and deflate Mutation}
\newacronym{ne}{NE}{Neuroevolution}
\newacronym{ngspt}{NEVO-GSPT}{NeuroEVOlution through Geometric Semantic perturbation and Population based Training}
\newacronym{mlp}{MLP}{Multi-layer Perceptron}
\newacronym{slm}{SLM}{Semantic Learning Machine}
\newacronym{nn}{NN}{Neural Network}
\newacronym{nns}{NNs}{Neural Networks}
\input{gls_dict}

\begin{document}

\title{NEVO-GSPT: Population-Based Neural Network Evolution Using Inflate and Deflate Operators}
%
\titlerunning{Population-Based Evolution of ANNs Using Inflate and Deflate Operators}

%

\author{Davide Farinati\inst{1,3} \and Frederico JBB Santos\inst{2} \and Leonardo Vanneschi\inst{1} \and Mauro Castelli\inst{1}}
\authorrunning{D. Farinati et al.}
%
\institute{NOVA Information Management School (NOVA IMS), Universidade Nova de Lisboa, Campus de Campolide, 1070-312 Lisboa, Portugal \\
\email{\{dfarinati, lvanneschi, mcastelli\}@novaims.unl.pt}
\and
Department of Engineering and Architecture, University of Trieste, Italy. \\
\email{fredericojose.jacomedebritosantos@phd.units.it}
\and
Vita-Salute San Raffaele University, Milan, Italy
Comprehensive Cancer Center/Unit of Urology; URI; IRCCS Ospedale San Raffaele, Milan, Italy
 \\
\email{farinati.davide@hsr.it}
}

\maketitle              

\begin{abstract}
Evolving neural network architectures is a computationally demanding process. Traditional methods often require an extensive search through large architectural spaces and offer limited understanding of how structural modifications influence model behavior.
This paper introduces \gls{ngspt}, a novel Neuroevolution algorithm based on two key innovations. 
First, we adapt geometric semantic operators~(GSOs) from genetic programming to neural network evolution, ensuring that architectural changes produce predictable effects on network semantics within a unimodal error surface.
Second, we introduce a novel operator (DGSM) that enables controlled reduction of network size, while maintaining the semantic  properties of~GSOs.
Unlike traditional approaches, \gls{ngspt}'s efficient evaluation mechanism, which only requires computing the semantics of newly added components, allows for efficient population-based training, resulting in a comprehensive exploration of the search space at a fraction of the computational cost.
Experimental results on four regression benchmarks show that \gls{ngspt} consistently evolves compact neural networks that achieve performance comparable to or better than established methods in the literature, such as standard neural networks, SLIM-GSGP, TensorNEAT, and SLM.

\keywords{Neuroevolution, Evolutionary Machine Learning, Artificial Neural Networks, Population Based Training, Geometric
Semantic
Mutation}
\end{abstract}
%
%



\section{Introduction}
\label{sec: intro}

The design of effective \gls{anns} architecture
represents one of the core challenges in modern machine learning. Despite \gls{anns}' success across numerous domains, determining the optimal architecture for a given task typically requires extensive manual experimentation or computationally costly automated search~\cite{gho22}. Traditional optimization methods such as grid and random search must evaluate each architecture from scratch and offer no principled link between architectural modifications and their effects on network behavior, making them both inefficient and opaque.
\gls{nas}~\cite{el19} automates this process but at a heavy computational price, often consuming thousands of GPU-hours and thus remaining accessible only to few institutions. Moreover, \gls{nas} approaches typically treat architecture optimization as a black-box problem, providing little interpretability or understanding of why certain designs outperform others. 
\gls{ne}~\cite{risi25}, which uses evolutionary algorithms to optimize \gls{anns}, offers a compelling alternative. Early \gls{ne} methods, such as \gls{neat}~\cite{stanley2002evolving}, successfully evolved both topologies and weights through principled genetic representations, but they scale poorly to modern deep architectures and remain computationally expensive because each individual must be fully trained or evaluated. More recent \gls{ne} variants~\cite{cu25,hu25} have attempted to mitigate these costs, yet the central challenge persists: efficiently exploring the architecture space while maintaining predictable relationships between structure and behavior.

To address this, the present work introduces NeuroEVOlution through Geometric Semantic perturbation and Population based Training~(NEVO-GSPT), a novel \gls{ne} method that enhances both computational efficiency and semantic awareness. \gls{ngspt} adapts to the evolution of \gls{anns} the basic principles of Geometric Semantic Genetic Programming (GSGP)~\cite{PaperMoraglio} and its recent development called Semantic Learning algorithm based on Inflate and deflate mutations~(SLIM-GSGP)~\cite{slim1,slim2}.
\gls{ngspt} extends a preexisting framework called \gls{slm}~\cite{gonccalves2015semantic}, using operators that have predictable geometric effects on the semantics of the individuals and induce a unimodal error surface 
for any supervised learning task.
One of the key innovations of~\gls{ngspt} compared to~\gls{slm} is the introduction of the Deflate Geometric Semantic Mutation~(DGSM). While the traditional Inflate operator~(IGSM) expands networks by adding perturbation components,
DGSM reduces complexity by removing previously added elements. Alternating between inflation and deflation prevents uncontrolled model growth, enabling the evolution of compact, interpretable networks while preserving the favorable properties of geometric semantic operators.
\gls{ngspt} is also computationally efficient, because, following the inspiration of SLIM-GSGP, it maintains networks as linked lists of perturbation components. When mutation occurs, only the new or removed component must be evaluated, while previous evaluations are reused. This incremental mechanism eliminates the need to retrain entire networks, drastically reducing fitness computation costs. 
Our experimental study, conducted on four regression problems, was designed to investigate four key factors: (1)~the~effect of a priori backpropagation training of the initial population; (2)~the~benefits of a posteriori fine-tuning of evolved models; (3)~the~impact of different inflate–deflate probabilities; and (4)~the~influence of the complexity of added components.

The manuscript is organized as follows. Section~\ref{sec:litrev} reviews related work in~\gls{ne}. Section~\ref{sec: back} provides background information allowing the reader to frame this work. Section~\ref{sec: meth} presents our methodology, detailing the~IGSM and~DGSM for~NE and the adopted population-based training approach. Section~\ref{sec: exp an} describes our experimental study. 
Finally, Section~\ref{sec: concl} concludes the paper, also discussing directions for future work.

\section{Related Work}
\label{sec:litrev}

\subsection{Neuroevolution Methods}
The field of \gls{ne} has evolved significantly since early methods like \gls{neat}~\cite{stanley2002evolving}, which allow evolving both network topology and weights through speciation and incremental complexity growth. Extensions such as HyperNEAT~\cite{stanley2009hypercube} and ES-HyperNEAT~\cite{risi2012unified} employed indirect encodings via CPPNs (Compositional Pattern Producing Networks) to evolve large-scale networks with regularities, though primarily for shallow architectures. 
More recent approaches have focused on deep architectures: CoDeepNEAT~\cite{miikkulainen2018evolving} evolves deep neural networks by introducing a two-level evolutionary process, where one population evolves modules (groups of layers), while another population evolves how these modules are assembled into complete networks; CGP-CNN (Cartesian Genetic Programming for Convolutional Neural Networks)~\cite{suganuma2017genetic,suganuma2020evolution} uses Cartesian Genetic Programming to represent CNNs as directed acyclic graphs; methods like DENSER~\cite{assunccao2019denser} employ dynamic structured grammatical evolution. While these approaches have demonstrated competitive results on benchmarks such as ImageNet and CIFAR, they face a common challenge: computational cost, often requiring hundreds to thousands of GPU-days~\cite{real2019regularized}. This computational barrier has limited their practical application and accessibility.

\subsection{Semantic-Based Approaches}
A fundamentally different approach emerged from Geometric Semantic Genetic Programming (GSGP), which introduced operators with predictable effects on program semantics, inducing unimodal fitness landscapes. The \glsreset{slm}\gls{slm}~\cite{gonccalves2015semantic} adapted these principles to neural network construction, incrementally adding neurons through geometric semantic mutation with constrained output contributions.
However, \gls{slm} has limitations: \begin{enumerate*}[label=(\arabic*)]
    \item it does not use gradient-based optimization (backpropagation), instead relying solely on evolution for parameter optimization; and
    \item the evolved networks, while more compact than traditional \gls{gsgp} trees, can still grow substantially.
\end{enumerate*}
SLIM~-~GSGP~\cite{pietropolli2025introducing} addressed the growth problem by introducing deflate mutation alongside inflate mutation, allowing both expansion and reduction of program size while maintaining the unimodal fitness landscape property. This approach has shown promising results in reducing model size compared to standard \gls{gsgp}, though it has primarily been applied to symbolic regression rather than neural network evolution.

All in all, the need for a \gls{ne} method that can efficiently evolve compact, interpretable neural network architectures while maintaining competitive performance on modern benchmarks remains a significant gap in the literature~\cite{white2023neural}. 
Such a method would need to: \begin{enumerate*}[label=(\arabic*)]
    \item leverage gradient-based optimization for parameter learning while evolving topology;
    \item maintain strict control over the size of the individuals;
    \item induce favorable fitness landscape properties to facilitate efficient search; and
    \item achieve practical training times measured in GPU-minutes (rather than GPU-days).
\end{enumerate*}
The method we propose in this paper was designed taking these four guidelines into account.

\section{Background}
\label{sec: back}

\subsection{Geometric Semantic Genetic Programming}
\label{sec: gsgp}

In supervised learning tasks, \gls{gp} individuals are generally represented as computer programs that translate input values into output results. For a given individual~$P$ and a dataset of $n$ fitness cases $X = \{x_{1}, x_{2}, \dots, x_{n}\}$, the resulting output vector \mbox{$s_{P} = \{ P(x_{1}), P(x_{2}), \dots, P(x_{n})\}$} is known as the individual's \emph{semantics}~\cite{PaperMoraglio}.
Traditional \gls{gp} operators like crossover and mutation modify program structure syntactically, making their semantic effects difficult to predict and potentially disrupting search locality~\cite{Nguyen:thesis}.

\gls{gsgp}~\cite{PaperMoraglio} addresses this by replacing syntactic operators with \gls{gso}s that enforce specific geometric behaviors in semantic space, inducing a unimodal error surface for any supervised learning task. For a parent function $T : \mathbb{R}^n \rightarrow \mathbb{R}$, \textit{Geometric Semantic Mutation} (GSM) defines a new function as $T_M = T + \text{ms} \cdot (T_{R1} - T_{R2})$, where ms denotes the mutation step and $T_{R1}$ and $T_{R2}$ are random functions whose outputs lie within $[0, 1]$. This ensures the mutated individual lies within a hypersphere of radius $ms$ centered at the parent's semantics.


Inspired by the success of \gls{gsgp}, various extensions have been proposed in recent years~\cite{pietropolli2023parametrizing,castelli2015geometric,bonin2024cellular}. However, a significant drawback of the method is that offspring tend to be larger than their parents, leading to solutions that are often difficult to interpret~\cite{PaperMoraglio}.


\subsection{\acrlong{slim}}
\label{sec: slim}

To address GSGP's bloating problem, recent work~\cite{slim1} introduced \gls{dgsm}, which reduces individual size while preserving semantic properties. The newly proposed operator builds on two core ideas. First, the~\gls{igsm} from the previous section can be algebraically reformulated as:
\[
\gls{igsm}\left(T\right) = T + \text{ms} \cdot \left(T_{R1} - T_{R2}\right) = T - \text{ms} \cdot \left(T_{R2} - T_{R1}\right),
\]
based on the identity $ \left(T_{R1} - T_{R2}\right) = - \left(T_{R2} - T_{R1}\right) $. Second, since $T_{R1}$ and $T_{R2}$ are randomly generated from the same distribution, they are interchangeable. Thus, another valid expression for~\gls{igsm} is:
\[
\gls{igsm}\left(T\right) = T - \text{ms} \cdot \left(T_{R1} - T_{R2}\right).
\]

Consider an individual $T$ that has undergone three applications of~\gls{igsm}, resulting in:
\[
T + \text{ms} \cdot \left(T_{R1} - T_{R2}\right) + \text{ms} \cdot \left(T_{R3} - T_{R4}\right) + \text{ms} \cdot \left(T_{R5} - T_{R6}\right).
\]
Using the alternative subtraction-based formulation of~\gls{igsm}, we can apply a fourth mutation:
\[
T + \text{ms} \cdot \left(T_{R1} - T_{R2}\right) + \text{ms} \cdot \left(T_{R3} - T_{R4}\right) + \text{ms} \cdot \left(T_{R5} - T_{R6}\right) - \text{ms} \cdot \left(T_{R7} - T_{R8}\right).
\]
This continues to expand the individual. However, by reusing an earlier pair of random programs (such as $T_{R3}$ and $T_{R4}$) the individual can be simplified:
\[
T + \text{ms} \cdot \left(T_{R1} - T_{R2}\right) \cancel{+ \text{ms} \cdot \left(T_{R3} - T_{R4}\right)} + \text{ms} \cdot \left(T_{R5} - T_{R6}\right) \cancel{- \text{ms} \cdot \left(T_{R3} - T_{R4}\right)}.
\]

The middle terms cancel, yielding a more compact individual. DGSM implements this by removing previously added terms while maintaining all semantic properties of GSM, including bounded perturbations and unimodal error surfaces. As the name implies, the~\gls{slim} algorithm incorporates both~\gls{igsm} and~\gls{dgsm} with configurable probabilities, restricting DGSM when only one term remains.

\subsection{\acrlong{slm}}
\label{sec: slm}
The \gls{slm}~\cite{gonccalves2015semantic} applied \gls{gsgp} principles to feedforward neural networks. 
Specifically, when mutating a parent network $T$, the \gls{slm} creates an offspring by adding a randomly generated network whose output is scaled by $ms$, ensuring that the semantic change remains bounded within a sphere of radius $ms$ around the parent's semantics. This approach guarantees that offspring remain within a controlled region of the semantic space, similar to the ball mutation property of \gls{gsm} in \gls{gsgp}.
\gls{slm}'s limitations include: \begin{enumerate*}[label=(\arabic*)]
    \item no backpropagation for parameter optimization (the parameters of the random networks are initialized and remain fixed), relying solely on evolution for improving network performance; and 
    \item potential for substantial network growth despite being more compact than \gls{gsgp} trees.
\end{enumerate*}

\subsection{\acrlong{tneat}}
\label{sec:tensorneat}
\gls{tneat}~\cite{wang2024tensorized} addresses a computational bottleneck in \gls{ne}: the inability to efficiently parallelize \gls{neat}'s operations across GPU architectures. 
\gls{tneat} overcomes this limitation by tensorizing networks with diverse topologies into uniform tensors for batched GPU computation. Networks are encoded as fixed-size tensors with padding, allowing all \gls{neat} operations (mutation, crossover, speciation, and network inference) to be vectorized across the entire population and to be executed in parallel on GPUs. 
\gls{tneat} achieves significant speedups over traditional implementations. Experimental results demonstrate speedups exceeding 500\( \times \) compared to \verb|NEAT-Python|, the most widely used open-source \gls{neat} implementation.

\section{Methodology}
\label{sec: meth}
In this work, we present \gls{ngspt}, a novel \gls{ne} approach that integrates the principles of \gls{gsgp} and \gls{slim} to evolve a population of \glspl{nn}.
This method simultaneously evolves the architecture, weights, and activation functions of the \glspl{nn} in the population.
The algorithm relies on two core operators: \gls{igsm} and \gls{dgsm}, both inspired by the \gls{gso} described in \Cref{sec: back}.

The objective of this work is to reproduce the effect of the operators used in \gls{gsgp} and \gls{slim} on \gls{nn}s.

\subsection{\acrlong{igsm} for \acrlong{ne}}
\label{sec: igsm nevo}

The \gls{igsm} is a \gls{ne} operator designed to introduce controlled variation into an existing neural network \( N \), while preserving its original functionality. 
This is achieved by appending a new auxiliary random network \( R \), which acts as a perturbation to the parent model.
The network \( R \) is constructed under the following conditions:
\begin{itemize}
    \item It contains exactly one neuron per layer of \( N \), totaling \( n \) neurons for an \( n \)-layer network.
    \item It receives connections only from the parent network; no backward connections into \( N \) are allowed.
    This ensures that \( N \)'s semantics remain unchanged.
    \item All internal weights in \( R \) are drawn uniformly from the range \( [-1, 1] \), except for the final output weight.
    \item The output weight is drawn from \( [0, ms] \), where \( ms \in \mathbb{R} \) is a tunable parameter known as the mutation step.
    \item The output neuron of \( R \) uses a \texttt{TanH} activation function to ensure bounded outputs in \( [-1, 1] \).
    \item Other neurons in \( R \) may use arbitrary activation functions.
\end{itemize}

Once constructed, the semantic contribution of \( R \) is scaled by the mutation step and added to the output of \( N \), producing the final offspring network. 
The \gls{igsm} guarantees that the overall effect of the perturbation lies within the range \( [-ms, ms] \), thereby offering a bounded, additive modification to the parent network's behavior.
This operation can be concisely expressed as:
\begin{equation}
    \text{\gls{igsm}}(N) = N + ms \cdot R
\label{eq: igsm nevo}
\end{equation}

\begin{figure}
    \centering
    \begin{subfigure}{0.45\textwidth}
    \centering
    \includegraphics[width = 0.9\linewidth]{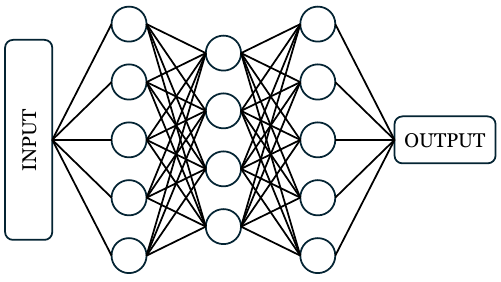}
    \caption{Parent \gls{nn}}
    \end{subfigure}
    \hfill
    \begin{subfigure}{0.45\textwidth}
    \centering
    \includegraphics[width = 0.5\linewidth]{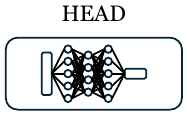}
    \caption{Parent \gls{nn} as a linked list}
    \end{subfigure}
    \vfill
    \begin{subfigure}{0.45\textwidth}
    \centering
    \includegraphics[width = \linewidth]{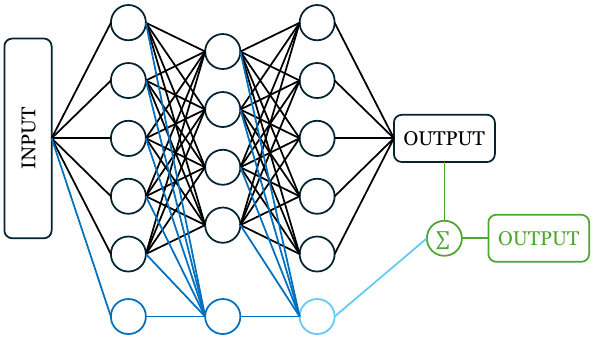}
    \caption{\gls{igsm} applied to the network}
    \end{subfigure}
    \hfill
    \begin{subfigure}{0.45\textwidth}
    \centering
    \includegraphics[width = 0.6\linewidth]{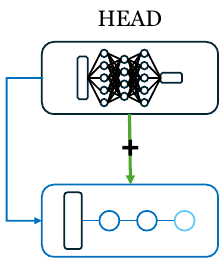}
    \caption{\gls{igsm} applied in linked list form}
    \end{subfigure}

    \caption{Illustration of \gls{igsm} using both neural network and linked list representations.}
    \label{fig:igsm}
\end{figure}

As with the \gls{slim} algorithm, the resulting structure of the offspring can be interpreted as a linked list. \Cref{fig:igsm} illustrates both the standard neural network representation and its linked list counterpart before and after applying the \gls{igsm} transformation.

\subsection{\acrlong{dgsm} for Neuroevolution}
\label{sec: dgsm nevo}

The \gls{dgsm} is a complementary operator to \gls{igsm} that introduces a mechanism for semantic simplification in evolved neural networks. While \gls{igsm} incrementally adds new perturbation components to the output of a neural network, \gls{dgsm} reverses this process by removing previously added perturbations. This allows for the reduction of model complexity without compromising semantic fidelity.

In the context of \gls{ne}, \gls{dgsm} is implemented by identifying a previously added auxiliary network \( R_i \) (introduced via \gls{igsm}) and subtracting its contribution from the current network. Given a neural network \( N \) that has undergone multiple \gls{igsm} transformations, its output can be expressed as:

\[
N' = N + ms \cdot R_1 + ms \cdot R_2 + \cdots + ms \cdot R_k
\]

\gls{dgsm} selects a random \( R_j \in \{R_1, R_2, \ldots, R_k\} \) and removes its contribution:

\[
\gls{dgsm}(N') = N' - ms \cdot R_j
\]

This operation preserves the core semantic properties of geometric semantic mutation: it induces bounded, linear perturbations to the network output and maintains the unimodal structure of the error landscape. 
However, it simultaneously counters the inherent bloating problem of repeated additive perturbations by reducing the effective size of the evolved model.

To enable this behavior, a history of perturbation components must be maintained in a linked list-like structure, where each element corresponds to a specific auxiliary network previously appended via \gls{igsm}. 
\gls{dgsm} acts by deleting an element from this list.

\begin{figure}[!ht]
    \centering
    \begin{subfigure}{0.45\textwidth}
    \centering
    \includegraphics[width = 0.9\linewidth]{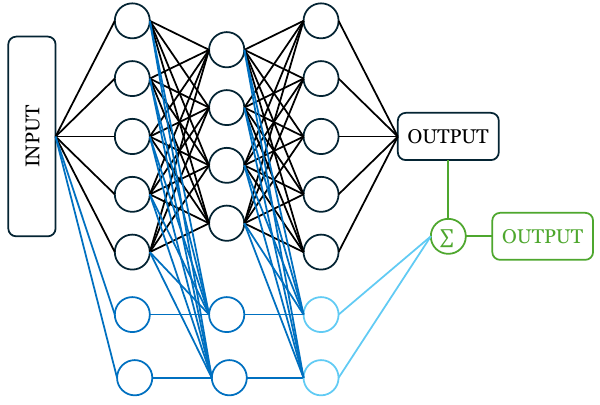}
    \caption{Parent \gls{nn} after undergoing two \gls{igsm}}
    \end{subfigure}
    \hfill
    \begin{subfigure}{0.45\textwidth}
    \centering
    \includegraphics[width = 0.35\linewidth]{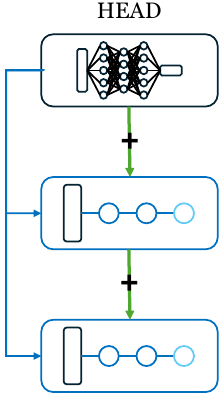}
    \caption{Parent \gls{nn} after undergoing two \gls{igsm} as a linked list}
    \end{subfigure}
    \vfill
    \begin{subfigure}{0.45\textwidth}
    \centering
    \includegraphics[width = 0.9\linewidth]{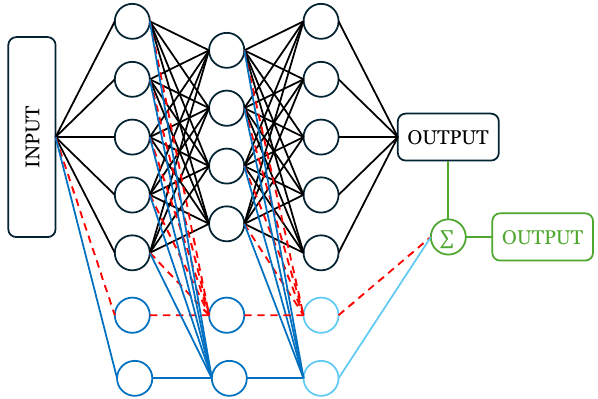}
    \caption{\gls{dgsm} applied to the network}
    \end{subfigure}
    \hfill
    \begin{subfigure}{0.45\textwidth}
    \centering
    \includegraphics[width = 0.35\linewidth]{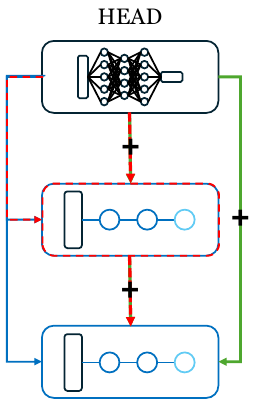}
    \caption{\gls{dgsm} applied in linked list form}
    \end{subfigure}

    \caption{Illustration of \gls{dgsm} using both neural network and linked list representations.}
    \label{fig:dgsm}
\end{figure}

Figure~\ref{fig:dgsm} illustrates the application of \gls{dgsm} in both neural network and linked list form.

\subsection{Population based training}
\label{sec: pbt}

The evaluation of new individuals is performed by summing the semantics of all elements in the linked list, as proposed in previous work~\cite{slim2}.
As a result, when \gls{igsm} is applied, only a single \gls{nn} with a width of one needs to be evaluated, since the semantics of all other components have already been computed.
In contrast, when \gls{dgsm} is used, the semantics and fitness of the new individual can be determined through a simple summation.
This makes the evaluation process highly efficient, enabling the evolution of a population of individuals.
Maintaining a population allows for a more comprehensive exploration of the fitness landscape associated with a given problem.

\section{Experimental Analysis}
\label{sec: exp an}

This section presents the experiments conducted to evaluate \gls{ngspt}. 
It is organized into three parts. \Cref{sec: exp sett} describes the main benchmarks used to assess the performance of the proposed algorithm. 
\Cref{sec: abl studies} provides an in-depth analysis of the various hyperparameters and configurations of \gls{ngspt}. 
Finally, \Cref{sec: exp res} reports the performance of \gls{ngspt} compared to other baseline algorithms from \gls{ec}, \gls{ne}, and \gls{ml}.

\subsection{Experimental Settings}
\label{sec: exp sett}
In order to evaluate \gls{ngspt} we chose four regression datasets commonly used as benchmark to evaluate \gls{gp}~\cite{Archetti2007,DBLP:conf/gecco/SilvaV09,cast13,chull,semanticNeighborhoodIvo,McDermott2012,pietropolli2023hybridization}, \gls{ne}~\cite{gonccalves2015semantic} and \gls{nn}~\cite{dong_liu_dai_2024}.
The used datasets and their characteristics are presented in \Cref{tab:data}.
These datasets offer different overfitting profiles~\cite{McDermott2012,farinati_ieee}, making it possible to study the generalization capabilities of \gls{ngspt}.

\begin{table}[!ht]
   
    \centering
    \begin{tabular}{|c|c|c|}
    \hline
        \textbf{Dataset} & \textbf{Number of observations} & \textbf{Number of features}  \\
        \hline
        \hline
         airfoil & 1502  & 5 \\
         \hline
         concrete strength & 1029 & 8\\
         \hline
         bioavailability & 359 & 241\\
         \hline
         ld50 & 234 & 626 \\
         \hline
    \end{tabular}
 \caption{Characteristics of the datasets used for model comparison.}
    \label{tab:data}
\end{table}

Several hyperparameters were analyzed, and their optimal values were determined in \Cref{sec: abl studies}. 
Other hyperparameters were kept fixed throughout all experiments, with the population size set to 100 and the mutation step fixed at 2.
For all comparisons, a 80-20 train–test split Monte Carlo cross-validation was employed, using consistent splits for each run across methods.
The experiments presented in \Cref{sec: abl studies} were conducted over ten independent runs, whereas those reported in \Cref{sec: exp res} were performed over thirty runs.
The corresponding p-values resulting from the pairwise Wilcoxon signed-rank test for the latter experiments are provided in the Supplementary Material.
The python code used for this work can be found at \hyperlink{https://github.com/Fari98/NEVO-GSPT}{\gls{ngspt}}\footnote{https://github.com/Fari98/NEVO-GSPT}.

\subsection{Ablation Studies}
\label{sec: abl studies}

This Section provides an in-depth analysis of the algorithm’s configurations and hyperparameters, focusing on identifying the optimal setup for each dataset and explaining the underlying reasons for its effectiveness.

\subsubsection{A priori training}
\label{sec: init}

During population initialization, the $|P|$ \glspl{nn} are randomly generated with varying structures and weights, where $|P|$ denotes the population size.
Each \gls{nn} can either be trained using backpropagation or directly passed to the evolutionary process.
This section investigates the impact of \emph{A priori Training}~(AprT) on the evolution process. For each dataset, three configurations are evaluated: no AprT, AprT applied to half of the \glspl{nn} (selected randomly), and AprT applied to all \glspl{nn}. 
The results, presenting the RMSE on the training and test sets for each dataset, are shown in \Cref{fig: pre train}.

\input{figures/init_technique}

The results in \Cref{fig: pre train} demonstrate that the Half AprT configuration consistently achieves the best balance between training and test RMSE across all datasets.
While No AprT exhibits slower convergence and higher error, and Full AprT tends to overfit in some cases, Half AprT provides faster convergence and superior generalization performance.
This indicates that a moderate level of AprT integration optimally guides the evolutionary process.

\subsubsection{A posteriori training}
\label{sec: post train}

The final solutions generated by \gls{ngspt} are \glspl{nn} that can also be further trained to optimize their weights.
This Section examines the effect of \emph{A posteriori Training}~(ApoT) on the final solutions obtained at the last generation.
The results, showing the test RMSE for each dataset, are presented in \Cref{fig:bp apot}.

\input{figures/post_train_bp}

As shown in the results, ApoT has a positive effect on datasets that are less prone to overfitting (i.e., \textit{concrete\_strength} and \textit{airfoil}), improving the test RMSE. 
However, the opposite trend is observed in datasets susceptible to overfitting (i.e., \textit{bioav} and \textit{ld50}), where ApoT worsens test performance and reduces the generalization capability of the algorithm.

\subsubsection{Deflate and Inflate probabilities}
\label{sec: prob}

Different combinations of inflate and deflate probabilities can be used to balance the trade-off between convergence speed and program size.
In this experiment, we tested four inflate probability values (0.3, 0.5, 0.7, and 1), with the corresponding deflate probability always set to one minus the inflate probability. 
The results of this analysis are shown in \Cref{fig: prob} and \Cref{fig: size}, presenting the RMSE performance on the training and test sets, as well as the resulting program sizes.

\input{figures/probabilities_perf}
\input{figures/probabilities_size}

As expected, lower inflate probabilities produce smaller programs. 
However, the impact on performance is less straightforward. 
For datasets that are not prone to overfitting (i.e., \textit{concrete\_strength} and \textit{airfoil}), a low inflate probability results in slower convergence. 
Conversely, for datasets prone to overfitting (i.e., \textit{bioav} and \textit{ld50}), a lower inflate probability improves the algorithm’s generalization capability.
These results reinforce the observation reported in~\cite{farinati_ieee} that, for such problems, program size tends to correlate with overfitting.

\subsubsection{Neurons to be added}
\label{sec: neurons}

As explained in \Cref{sec: igsm nevo}, when the \gls{igsm} operator is applied, one neuron per layer is added to the parent \gls{nn}.
However, more complex or simpler networks could also be generated. 
In this section, we investigate the impact of adding fewer neurons per layer by testing different proportions as hyperparameters: 0.3, 0.5, 0.7, and 1.
\Cref{fig: neurons} shows the training and test RMSE performance for each of these configurations.

\input{figures/n_neurons}

As observed in \Cref{fig: neurons}, adding fewer than one neuron per layer consistently results in slower convergence and reduced generalization capability of the algorithm.

\subsection{Experimental Results}
\label{sec: exp res}

After exploring the hyperparameters of \gls{ngspt} and selecting the best configurations for each dataset, distinguishing clearly between those prone to overfitting and those that are not, we compare the algorithm against a set of baselines from \gls{ec} and \gls{ne}. 
Among the baselines, we include a \gls{nn} with the same architecture as the one evolved by \gls{ngspt}, whose weights are optimized via backpropagation, as well as \gls{slm}~\cite{gonccalves2015semantic}, \gls{tneat}~\cite{wang2024tensorized}, and SLIM-GSGP~\cite{rosenfeld2025slimgsgp} (variant \textit{SLIM*SIG1}).
For fair comparison, all evolutionary algorithms are configured with consistent hyperparameters to match the number of parameters of an evolved \gls{ngspt} model.
\gls{tneat} is run for 1000 generations with a population size of 100, using a maximum of 627 nodes and 626 connections, with 10 individuals per species and a survival threshold of 0.2. For the initial population of \gls{tneat}, there must be at least as many connections and nodes as input features, hence the higher values when compared to \gls{ngspt}.
\gls{slm} is executed for 100 generations using networks with a maximum of 3 layers, starting with 32 initial neurons and adding 4 neurons per layer per \gls{gsm} mutation. Note the shift in increasing the number of neurons per mutations and decreasing the number of generations is necessary due to limitations in \gls{slm}'s implementation (PyTorch Genesis~\cite{santos2023neuroevolution}) creating significant computational overhead when adding only neuron per layer per generation. We therefore adjust the parameters to ensure a fair comparison.
\Cref{fig:bp} presents the test RMSE of the different algorithms across all datasets. 
The p-values of the Wilcoxon signed-rank test are provided in the Supplementary Material.

\input{figures/final_results_bp}

As observed in \Cref{fig:bp}, \gls{ngspt} is the only algorithm that consistently ranks among the top performers across all tasks.
This highlights the effectiveness of population-based training combined with geometric mutation in achieving both strong convergence and robust generalization.

\subsubsection{Comparison of program size}
\label{sec: comp size}

The novelty of this work also lies in the introduction of a \gls{dgsm} operator for \gls{ne}, designed to generate more compact \glspl{nn}. 
\Cref{fig:bp size} compares the number of neurons produced by \gls{ngspt} with those obtained by two other \gls{ne} techniques, namely \gls{tneat} and \gls{slm}, across all datasets.

\input{figures/size}

As shown in all test cases, \gls{ngspt} consistently produces smaller \glspl{nn} than \gls{slm}, demonstrating the effectiveness of the proposed \gls{dgsm} operator. 
Moreover, in two out of four datasets, \gls{ngspt} also generates smaller networks than \gls{tneat}, a state-of-the-art algorithm in \gls{ne}.

\subsubsection{Empirical Analysis of the running time}
\label{sec: run time}

To assess the computational efficiency of NEVO-GSPT, we conducted 30 independent runs comparing runtime performance against standard backpropagation on an Intel Core i5-1235U (1.30 GHz, 16GB RAM). 
Over 100 generations/epochs, NEVO-GSPT averaged 14.58s ± 0.60s while backpropagation required 5.12s ± 0.42s. 
However, this comparison requires important contextualization: each NEVO-GSPT generation evaluates 100 individuals in the population, whereas backpropagation represents a single model training run.
When examining individual evaluation time, NEVO-GSPT's inflate/deflate operations complete in 0.0006s ± 0.00008s compared to 0.0031s ± 0.0006s for backpropagation, demonstrating approximately 5× faster individual evaluations.
It is worth noting that direct runtime comparisons with GPU-accelerated implementations such as PyTorch Genesis (used by SLM) and TensorNEAT may not be entirely equitable, as our CPU-based implementation operates under different computational constraints.
Nevertheless, NEVO-GSPT's ability to explore 100 architectural solutions simultaneously within approximately 15 seconds on CPU hardware represents an advancement for \gls{ne}, particularly in environments where GPU access may be limited.

\section{Conclusions}
\label{sec: concl}

This paper introduced \gls{ngspt}, a novel \gls{ne} algorithm that addresses two fundamental challenges in neural architecture search: the computational expense of exploring architectural spaces and the lack of clear mappings between structural changes and their effects on network behavior. 
%
%
A key innovation of our work is the introduction of \gls{dgsm} for \gls{ne}. 
This novel operator enables controlled reduction of network complexity by removing previously added components, addressing the bloating problem that affects geometric semantic approaches while maintaining the favorable fitness landscape properties that facilitate efficient search. 
%
The computational efficiency of \gls{ngspt} represents another significant contribution. By maintaining networks as linked lists of perturbation components and evaluating only newly added or removed components, our method reduces the cost of fitness evaluation compared to approaches that must train entire networks from scratch. 
This efficiency makes it possible evolving large populations at a computational cost measured in minutes rather than GPU-days.

Experimental results across four regression benchmarks demonstrate that \gls{ngspt} consistently ranks among the top performers when compared against standard neural networks, \gls{slm}, \gls{tneat}, and SLIM-GSGP.
This competitive performance is achieved while producing significantly more compact networks than \gls{slm} and, in several cases, more compact than \gls{tneat}. 

\section*{Acknowledgments}

This work was supported by national funds through FCT (Fundação para a Ciência e a Tecnologia), under the project - \hyperlink{https://doi.org/10.54499/UID/04152/2025}{UID/04152/2025 - Centro de Investigação em Gestão de Informação (MagIC)/NOVA IMS}\footnote{https://doi.org/10.54499/UID/04152/2025} - (2025-01-01/2028-12-31) and \hyperlink{https://doi.org/10.54499/UID/PRR/04152/2025}{UID/PRR/04152/2025}\footnote{https://doi.org/10.54499/UID/PRR/04152/2025} (2025-01-01/ 2026-06-30).


\newpage

\bibliographystyle{splncs04}

\input{bibl.bbl}

\end{document}


The pseudo code of the \gls{ngspt} algorithm is presented in \Cref{algo: pseudocode}. 

\begin{algorithm}
\caption{\gls{ngspt} Algorithm}
\label{algo: pseudocode}
\begin{algorithmic}[1]
\State Initialize population $P$ with random \glspl{nn}
\State Evaluate fitness and semantics of all individuals in $P$
\While{termination condition not met}
        \If{elitism == \textbf{False}}
             \State $P' \gets \emptyset$
        \Else
            \State $P' \gets \text{elite}_P$
        \EndIf  
    \While{$|P'| < |P|$}
        \State Select parent $N$ from $P$ using a selection algorithm
        \If{random() $< p_{dgsm}$}
            \State $N' \gets$ apply \gls{dgsm} to $N$
        \Else
            \State $N' \gets$ apply \gls{igsm} to $N$
        \EndIf
       
        \State Add $N'$ to $P'$
    \EndWhile
    \State Evaluate semantics and fitness of all individuals in $P'$
    \State $P \gets$  $P'$ 
\EndWhile
\State \Return best individual in $P$
\end{algorithmic}
\end{algorithm}

\begin{table}
\caption{Statistical test comparing the test RMSE in the bioav dataset}
\begin{tabular}{lrllll}
\toprule
 & NEVO-GSPT & NN & TNEAT & SLIM & SLM \\
\midrule
NEVO-GSPT & NaN & 4.66e-03 & 1.86e-09 & 3.24e-06 & 2.35e-06 \\
NN & NaN & NaN & 1.86e-09 & 0.78 & 0.28 \\
TNEAT & NaN & NaN & NaN & 1.86e-09 & 1.86e-09 \\
SLIM & NaN & NaN & NaN & NaN & 0.4 \\
SLM & NaN & NaN & NaN & NaN & NaN \\
\bottomrule
\end{tabular}
\end{table}

\begin{table}
\caption{Statistical test comparing the test RMSE in the ld50 dataset}
\begin{tabular}{lrrlll}
\toprule
 & NEVO-GSPT & NN & TNEAT & SLIM & SLM \\
\midrule
NEVO-GSPT & NaN & 0.98 & 5.72e-07 & 0.31 & 0.57 \\
NN & NaN & NaN & 7.99e-06 & 0.97 & 0.38 \\
TNEAT & NaN & NaN & NaN & 8.01e-08 & 5.97e-06 \\
SLIM & NaN & NaN & NaN & NaN & 0.73 \\
SLM & NaN & NaN & NaN & NaN & NaN \\
\bottomrule
\end{tabular}
\end{table}

\begin{table}
\caption{Statistical test comparing the test RMSE in the concrete\_strength dataset}
\begin{tabular}{lrllll}
\toprule
 & NEVO-GSPT & NN & TNEAT & SLIM & SLM \\
\midrule
NEVO-GSPT & NaN & 1.86e-09 & 0.94 & 1.86e-09 & 1.86e-09 \\
NN & NaN & NaN & 1.86e-09 & 3.73e-09 & 5.38e-03 \\
TNEAT & NaN & NaN & NaN & 1.86e-09 & 1.86e-09 \\
SLIM & NaN & NaN & NaN & NaN & 3.73e-09 \\
SLM & NaN & NaN & NaN & NaN & NaN \\
\bottomrule
\end{tabular}
\end{table}

\begin{table}
\caption{Statistical test comparing the test RMSE in the airfoil dataset}
\begin{tabular}{lrllll}
\toprule
 & NEVO-GSPT & NN & TNEAT & SLIM & SLM \\
\midrule
NEVO-GSPT & NaN & 1.86e-09 & 1.13e-03 & 1.86e-09 & 1.86e-09 \\
NN & NaN & NaN & 1.86e-09 & 1.72e-03 & 3.80e-04 \\
TNEAT & NaN & NaN & NaN & 1.86e-09 & 1.86e-09 \\
SLIM & NaN & NaN & NaN & NaN & 0.9 \\
SLM & NaN & NaN & NaN & NaN & NaN \\
\bottomrule
\end{tabular}
\end{table}

\begin{table}
\caption{Statistical test comparing the number of nodes in the bioav dataset}
\begin{tabular}{lrll}
\toprule
 & NEVO-GSPT & TNEAT & SLM \\
\midrule
NEVO-GSPT & NaN & 1.86e-09 & 1.86e-09 \\
TNEAT & NaN & NaN & 1.86e-09 \\
SLM & NaN & NaN & NaN \\
\bottomrule
\end{tabular}
\end{table}

\begin{table}
\caption{Statistical test comparing the number of nodes in the ld50 dataset}
\begin{tabular}{lrll}
\toprule
 & NEVO-GSPT & TNEAT & SLM \\
\midrule
NEVO-GSPT & NaN & 1.86e-09 & 1.73e-06 \\
TNEAT & NaN & NaN & 1.86e-09 \\
SLM & NaN & NaN & NaN \\
\bottomrule
\end{tabular}
\end{table}

\begin{table}
\caption{Statistical test comparing the number of nodes in the concrete\_strength dataset}
\begin{tabular}{lrll}
\toprule
 & NEVO-GSPT & TNEAT & SLM \\
\midrule
NEVO-GSPT & NaN & 1.73e-06 & 1.86e-09 \\
TNEAT & NaN & NaN & 1.86e-09 \\
SLM & NaN & NaN & NaN \\
\bottomrule
\end{tabular}
\end{table}

\begin{table}
\caption{Statistical test comparing the number of nodes in the airfoil dataset}
\begin{tabular}{lrll}
\toprule
 & NEVO-GSPT & TNEAT & SLM \\
\midrule
NEVO-GSPT & NaN & 1.73e-06 & 1.73e-06 \\
TNEAT & NaN & NaN & 1.73e-06 \\
SLM & NaN & NaN & NaN \\
\bottomrule
\end{tabular}
\end{table}

%% file: gls_dict.tex
\newacronym{tneat}{TensorNEAT}{Tensorized NeuroEvolution of Augmenting Topologies}
\newacronym{neat}{NEAT}{NeuroEvolution of Augmenting Topologies}
\newacronym{nas}{NAS}{Neural Architecture Search}
\newacronym{anns}{ANNs}{Artificial Neural Networks}

%% file: figures/init_technique.tex
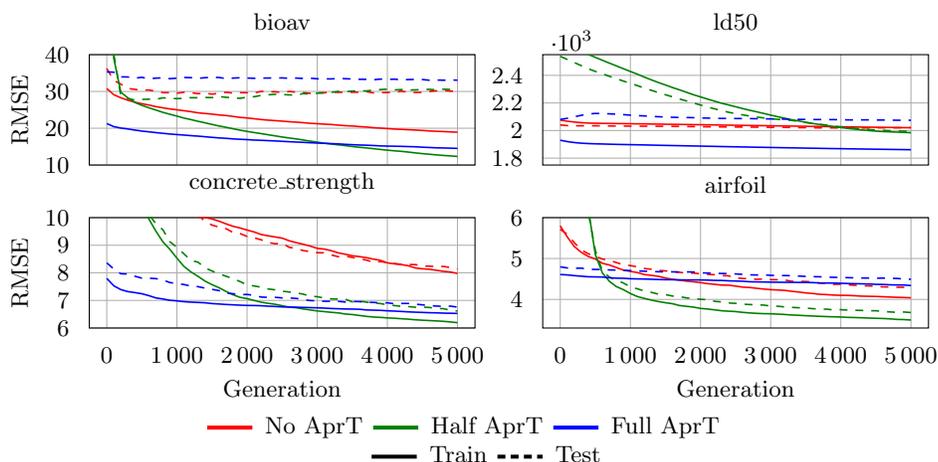
\begin{figure}[!h]
\centering
\begin{tikzpicture}

\definecolor{darkgray176}{RGB}{176,176,176}
\definecolor{green}{RGB}{0,128,0}
\definecolor{lightgray204}{RGB}{204,204,204}
\definecolor{purple}{RGB}{128,0,128}

\begin{groupplot}[ width=.55\textwidth,
            height=.25\linewidth,
            group style={group size=2 by 2,
                xticklabels at=edge bottom,
                horizontal sep=9mm,
                vertical sep=7mm,
                },
            grid=major,
            noinnerticks]
\nextgroupplot[
tick align=outside,
tick pos=left,
title={bioav},
x grid style={darkgray176},
xmin=-250, xmax=5250,
xtick style={color=black},
y grid style={darkgray176},
ylabel={RMSE},
ymin=10, ymax=40,
ytick style={color=black}
]
\addplot [semithick, red]
table {%
0 30.7764854431152
100 29.1160669326782
200 28.277322769165
300 27.6243858337402
400 27.086353302002
500 26.6545524597168
600 26.313268661499
700 25.8746814727783
800 25.5471754074097
900 25.245512008667
1000 25.0138053894043
1100 24.7638902664185
1200 24.5068845748901
1300 24.1909551620483
1400 24.014949798584
1500 23.8334503173828
1600 23.6184434890747
1700 23.4582567214966
1800 23.2139520645142
1900 22.9531698226929
2000 22.7876462936401
2100 22.5825328826904
2200 22.3671789169312
2300 22.2016525268555
2400 22.0690479278564
2500 21.9318294525146
2600 21.7734699249268
2700 21.6607551574707
2800 21.5112199783325
2900 21.3509483337402
3000 21.2319450378418
3100 21.1177787780762
3200 21.0147438049316
3300 20.845404624939
3400 20.700478553772
3500 20.6083631515503
3600 20.4430503845215
3700 20.3396673202515
3800 20.1488027572632
3900 20.0379867553711
4000 19.9084119796753
4100 19.7792768478394
4200 19.6941118240356
4300 19.5998315811157
4400 19.5116777420044
4500 19.3942174911499
4600 19.2625589370728
4700 19.175498008728
4800 19.0932340621948
4900 19.0074634552002
5000 18.9219369888306
};
\addplot [semithick, red, dashed]
table {%
0 36.2236633300781
100 33.0163803100586
200 31.9877042770386
300 30.812557220459
400 30.5633220672607
500 30.3364772796631
600 30.3883390426636
700 30.0649003982544
800 29.691704750061
900 29.6661024093628
1000 29.5137929916382
1100 29.7961797714233
1200 29.6628694534302
1300 29.4037456512451
1400 29.4929857254028
1500 29.2828903198242
1600 29.4631185531616
1700 29.7784156799316
1800 29.553692817688
1900 29.684100151062
2000 29.6548309326172
2100 29.5808868408203
2200 30.0225791931152
2300 29.9239664077759
2400 29.9014234542847
2500 29.7081871032715
2600 29.9868726730347
2700 29.816556930542
2800 29.6936264038086
2900 29.4877414703369
3000 29.674015045166
3100 29.6518354415894
3200 29.8232660293579
3300 29.7486782073975
3400 29.8201189041138
3500 29.7790088653564
3600 29.7996864318848
3700 29.9681119918823
3800 29.6139278411865
3900 29.6674318313599
4000 29.7158670425415
4100 29.7192430496216
4200 30.0799541473389
4300 29.7578973770142
4400 29.7402839660645
4500 29.7424125671387
4600 30.0001163482666
4700 30.171573638916
4800 30.2844285964966
4900 30.08385181427
5000 30.0976552963257
};
\addplot [semithick, green]
table {%
0 71.9247436523438
100 38.9549198150635
200 29.8104829788208
300 28.2554759979248
400 27.0116024017334
500 26.219349861145
600 25.5580377578735
700 24.9495258331299
800 24.3209505081177
900 23.7952337265015
1000 23.303750038147
1100 22.7606582641602
1200 22.3485736846924
1300 21.8897972106934
1400 21.4801387786865
1500 21.0400228500366
1600 20.61403465271
1700 20.2585783004761
1800 19.8601264953613
1900 19.473916053772
2000 19.1443958282471
2100 18.8279037475586
2200 18.5430822372437
2300 18.2313661575317
2400 17.9272394180298
2500 17.6227054595947
2600 17.3537755012512
2700 17.0986714363098
2800 16.8053846359253
2900 16.4876117706299
3000 16.2264604568481
3100 15.9287424087524
3200 15.7059025764465
3300 15.490954875946
3400 15.2782716751099
3500 15.0316224098206
3600 14.8243050575256
3700 14.6529579162598
3800 14.4497838020325
3900 14.235285282135
4000 14.0366144180298
4100 13.8672189712524
4200 13.7138109207153
4300 13.520103931427
4400 13.2957315444946
4500 13.0993323326111
4600 12.9278988838196
4700 12.7550692558289
4800 12.6065049171448
4900 12.4716029167175
5000 12.3364806175232
};
\addplot [semithick, green, dashed]
table {%
0 74.2934379577637
100 40.1591777801514
200 28.7833728790283
300 28.0976123809814
400 27.661958694458
500 27.8359041213989
600 27.8904809951782
700 27.8498821258545
800 28.2734546661377
900 28.1196441650391
1000 28.0403394699097
1100 28.1512880325317
1200 28.3050508499146
1300 28.28147315979
1400 28.3603048324585
1500 28.3340673446655
1600 28.61203956604
1700 28.4139356613159
1800 28.2642383575439
1900 28.1362438201904
2000 28.4364452362061
2100 28.5669889450073
2200 28.9198799133301
2300 29.0938282012939
2400 29.2247848510742
2500 29.1735534667969
2600 29.0651521682739
2700 29.0166969299316
2800 29.4267024993896
2900 29.2888717651367
3000 29.5257501602173
3100 29.5916213989258
3200 29.7500629425049
3300 30.0068864822388
3400 29.9342861175537
3500 30.0894289016724
3600 30.227126121521
3700 30.1321496963501
3800 30.3719081878662
3900 30.5120115280151
4000 30.6204471588135
4100 30.4123725891113
4200 30.516884803772
4300 30.3948640823364
4400 30.4286022186279
4500 30.5533361434937
4600 30.5633306503296
4700 30.4381141662598
4800 30.6107311248779
4900 30.5322036743164
5000 30.5449495315552
};
\addplot [semithick, blue]
table {%
0 21.2709989547729
100 20.4381170272827
200 20.0466642379761
300 19.8079223632812
400 19.4977254867554
500 19.2305040359497
600 19.0323610305786
700 18.8107509613037
800 18.6166086196899
900 18.4288377761841
1000 18.2801542282104
1100 18.145712852478
1200 17.9838275909424
1300 17.8596429824829
1400 17.6945133209229
1500 17.524564743042
1600 17.3715629577637
1700 17.2590961456299
1800 17.098705291748
1900 16.9979286193848
2000 16.9128484725952
2100 16.7797079086304
2200 16.719048500061
2300 16.6305828094482
2400 16.5456695556641
2500 16.4442138671875
2600 16.3381342887878
2700 16.2518939971924
2800 16.1627063751221
2900 16.0565633773804
3000 15.9730477333069
3100 15.8998627662659
3200 15.8193426132202
3300 15.7419180870056
3400 15.6197113990784
3500 15.5233097076416
3600 15.444477558136
3700 15.3897638320923
3800 15.3078641891479
3900 15.1981134414673
4000 15.1476769447327
4100 15.1091532707214
4200 15.0716376304626
4300 15.0143885612488
4400 14.9565906524658
4500 14.8430123329163
4600 14.7659592628479
4700 14.6739110946655
4800 14.6338243484497
4900 14.5861930847168
5000 14.5116901397705
};
\addplot [semithick, blue, dashed]
table {%
0 35.3149242401123
100 35.229829788208
200 33.964599609375
300 33.937894821167
400 33.6104431152344
500 33.8391571044922
600 33.4461574554443
700 33.6148242950439
800 33.710765838623
900 33.4656543731689
1000 33.6175403594971
1100 33.7973442077637
1200 33.7458438873291
1300 33.556791305542
1400 33.7371139526367
1500 33.8629131317139
1600 33.7899875640869
1700 33.5002288818359
1800 33.6250419616699
1900 33.7219486236572
2000 33.6463794708252
2100 33.5673408508301
2200 33.6204605102539
2300 33.6509246826172
2400 33.6638889312744
2500 33.6785144805908
2600 33.5758953094482
2700 33.6394462585449
2800 33.5268573760986
2900 33.6074028015137
3000 33.4331398010254
3100 33.4300727844238
3200 33.4015026092529
3300 33.5359401702881
3400 33.4588851928711
3500 33.3823661804199
3600 33.3026733398438
3700 33.2882328033447
3800 33.1649990081787
3900 33.364294052124
4000 33.285572052002
4100 33.3343257904053
4200 33.1749496459961
4300 33.2302932739258
4400 33.2645359039307
4500 33.2343807220459
4600 33.0485668182373
4700 33.1125659942627
4800 33.0478954315186
4900 33.122594833374
5000 33.0616455078125
};

\nextgroupplot[
tick align=outside,
tick pos=left,
title={ld50},
x grid style={darkgray176},
xmin=-250, xmax=5250,
xtick style={color=black},
y grid style={darkgray176},
scaled y ticks=base 10:-3,
ymin=1750, ymax=2550,
ytick style={color=black}
]
\addplot [semithick, red]
table {%
0 2075.82971191406
100 2067.62390136719
200 2061.66931152344
300 2057.07531738281
400 2054.11785888672
500 2052.87872314453
600 2052.07196044922
700 2051.42626953125
800 2050.63836669922
900 2049.97290039062
1000 2049.28668212891
1100 2048.56170654297
1200 2047.96783447266
1300 2047.18267822266
1400 2046.48767089844
1500 2045.64398193359
1600 2044.99664306641
1700 2044.41607666016
1800 2043.66137695312
1900 2042.85980224609
2000 2042.09893798828
2100 2041.36096191406
2200 2040.54748535156
2300 2039.57763671875
2400 2038.87579345703
2500 2038.23175048828
2600 2037.36065673828
2700 2036.66857910156
2800 2035.93249511719
2900 2035.36596679688
3000 2034.5322265625
3100 2033.77325439453
3200 2033.01483154297
3300 2032.22497558594
3400 2031.47094726562
3500 2030.68646240234
3600 2029.92028808594
3700 2029.14337158203
3800 2028.47503662109
3900 2027.76922607422
4000 2027.08148193359
4100 2026.33825683594
4200 2025.73779296875
4300 2024.98431396484
4400 2024.35833740234
4500 2023.65887451172
4600 2022.87774658203
4700 2022.30529785156
4800 2021.68188476562
4900 2020.96807861328
5000 2020.23162841797
};
\addplot [semithick, red, dashed]
table {%
0 2040.23083496094
100 2036.1767578125
200 2034.34735107422
300 2034.35522460938
400 2034.58709716797
500 2034.55389404297
600 2034.14514160156
700 2033.33605957031
800 2032.63848876953
900 2032.39831542969
1000 2031.95184326172
1100 2031.47967529297
1200 2031.36431884766
1300 2030.94610595703
1400 2031.01495361328
1500 2030.57885742188
1600 2030.22760009766
1700 2029.6220703125
1800 2028.94482421875
1900 2028.50500488281
2000 2027.64520263672
2100 2027.20397949219
2200 2026.69281005859
2300 2026.38037109375
2400 2026.05908203125
2500 2025.61083984375
2600 2025.03497314453
2700 2024.63983154297
2800 2024.47814941406
2900 2024.18200683594
3000 2023.25311279297
3100 2023.17510986328
3200 2022.80310058594
3300 2022.09704589844
3400 2020.89709472656
3500 2020.82373046875
3600 2019.84625244141
3700 2019.30987548828
3800 2019.78619384766
3900 2019.24835205078
4000 2018.3974609375
4100 2018.78570556641
4200 2018.77630615234
4300 2018.58770751953
4400 2019.37841796875
4500 2019.78765869141
4600 2019.79840087891
4700 2019.61505126953
4800 2019.98858642578
4900 2020.96472167969
5000 2021.96331787109
};

\addplot [semithick, green]
table {%
0 2642.01623535156
100 2617.22119140625
200 2594.19250488281
300 2572.66650390625
400 2551.19494628906
500 2530.53466796875
600 2509.33984375
700 2488.67041015625
800 2468.73400878906
900 2448.02624511719
1000 2427.98583984375
1100 2408.43322753906
1200 2387.14782714844
1300 2367.02038574219
1400 2347.90539550781
1500 2329.96862792969
1600 2311.65588378906
1700 2294.8828125
1800 2277.21533203125
1900 2259.36975097656
2000 2243.71130371094
2100 2227.93701171875
2200 2212.74035644531
2300 2197.95788574219
2400 2185.41784667969
2500 2172.58447265625
2600 2159.51232910156
2700 2147.04333496094
2800 2135.77868652344
2900 2123.98486328125
3000 2111.79699707031
3100 2100.1669921875
3200 2089.49291992188
3300 2078.60278320312
3400 2069.35485839844
3500 2059.56579589844
3600 2051.30737304688
3700 2043.39886474609
3800 2035.44006347656
3900 2028.08135986328
4000 2021.44067382812
4100 2015.62506103516
4200 2009.99200439453
4300 2004.57830810547
4400 1999.63909912109
4500 1995.51232910156
4600 1992.57727050781
4700 1990.50085449219
4800 1988.35626220703
4900 1986.27008056641
5000 1984.30456542969
};
\addplot [semithick, green, dashed]
table {%
0 2537.72229003906
100 2515.51013183594
200 2492.92907714844
300 2469.99658203125
400 2447.32763671875
500 2428.75500488281
600 2411.52307128906
700 2393.06555175781
800 2375.93908691406
900 2358.4189453125
1000 2341.89660644531
1100 2325.43774414062
1200 2307.90319824219
1300 2291.38134765625
1400 2274.94641113281
1500 2259.53479003906
1600 2244.23217773438
1700 2229.1328125
1800 2214.22375488281
1900 2199.09350585938
2000 2185.0703125
2100 2170.67138671875
2200 2157.33032226562
2300 2144.42822265625
2400 2132.43872070312
2500 2122.61206054688
2600 2113.44519042969
2700 2105.40417480469
2800 2098.08471679688
2900 2091.375
3000 2085.93640136719
3100 2081.171875
3200 2077.30859375
3300 2073.25231933594
3400 2066.57775878906
3500 2058.29040527344
3600 2051.5654296875
3700 2045.16577148438
3800 2039.51092529297
3900 2032.75823974609
4000 2026.13195800781
4100 2020.640625
4200 2016.63018798828
4300 2011.52423095703
4400 2007.34313964844
4500 2004.41192626953
4600 2001.43835449219
4700 1998.71520996094
4800 1997.00177001953
4900 1995.30035400391
5000 1994.37481689453
};
\addplot [semithick, blue]
table {%
0 1929.63519287109
100 1920.05603027344
200 1912.94750976562
300 1908.36785888672
400 1905.51391601562
500 1903.7919921875
600 1902.50659179688
700 1901.02996826172
800 1899.67224121094
900 1898.71008300781
1000 1897.72326660156
1100 1896.71569824219
1200 1895.63397216797
1300 1894.45581054688
1400 1893.26055908203
1500 1892.12377929688
1600 1891.15594482422
1700 1890.24340820312
1800 1889.13238525391
1900 1888.12548828125
2000 1887.15222167969
2100 1886.06274414062
2200 1885.10144042969
2300 1883.98577880859
2400 1882.9345703125
2500 1881.98876953125
2600 1880.99151611328
2700 1879.98046875
2800 1879.02728271484
2900 1878.14459228516
3000 1877.24652099609
3100 1876.24932861328
3200 1875.31585693359
3300 1874.59136962891
3400 1873.55712890625
3500 1872.6572265625
3600 1871.73059082031
3700 1870.94671630859
3800 1869.97186279297
3900 1869.18420410156
4000 1868.4755859375
4100 1867.64013671875
4200 1866.83227539062
4300 1866.11688232422
4400 1865.29998779297
4500 1864.62603759766
4600 1863.88232421875
4700 1863.11804199219
4800 1862.45135498047
4900 1861.880859375
5000 1860.99468994141
};
\addplot [semithick, blue, dashed]
table {%
0 2080.08911132812
100 2088.78918457031
200 2099.58435058594
300 2110.43139648438
400 2120.18823242188
500 2123.97326660156
600 2123.25549316406
700 2121.06494140625
800 2118.47717285156
900 2115.15991210938
1000 2111.69018554688
1100 2108.26843261719
1200 2105.89685058594
1300 2103.04040527344
1400 2100.08447265625
1500 2098.16003417969
1600 2096.17065429688
1700 2094.75988769531
1800 2094.38696289062
1900 2093.06799316406
2000 2091.18933105469
2100 2090.224609375
2200 2089.09753417969
2300 2087.85314941406
2400 2087.76403808594
2500 2087.09521484375
2600 2086.76599121094
2700 2086.61047363281
2800 2085.63159179688
2900 2084.64270019531
3000 2084.41052246094
3100 2083.95788574219
3200 2082.75573730469
3300 2081.25024414062
3400 2080.81604003906
3500 2080.50354003906
3600 2079.78918457031
3700 2079.83081054688
3800 2079.37390136719
3900 2079.24438476562
4000 2078.28649902344
4100 2077.27062988281
4200 2076.8466796875
4300 2077.4296875
4400 2077.23583984375
4500 2076.02331542969
4600 2076.34045410156
4700 2075.4033203125
4800 2074.71118164062
4900 2074.81872558594
5000 2074.60241699219
};

\nextgroupplot[
tick align=outside,
tick pos=left,
title={concrete\_strength},
x grid style={darkgray176},
xlabel={Generation},
xmin=-250, xmax=5250,
xtick style={color=black},
y grid style={darkgray176},
ylabel={RMSE},
ymin=6, ymax=10,
ytick style={color=black}
]
\addplot [semithick, red]
table {%
0 15.8872752189636
100 14.5600085258484
200 13.7327756881714
300 13.2490210533142
400 12.5470695495605
500 11.9741959571838
600 11.6842160224915
700 11.540198802948
800 11.3775253295898
900 11.0740962028503
1000 10.7568831443787
1100 10.4878978729248
1200 10.3229556083679
1300 10.1481566429138
1400 10.005717754364
1500 9.95090532302856
1600 9.8721604347229
1700 9.77474451065063
1800 9.68638944625854
1900 9.62556314468384
2000 9.55024528503418
2100 9.48160123825073
2200 9.38947916030884
2300 9.34204626083374
2400 9.29958963394165
2500 9.25788497924805
2600 9.16143846511841
2700 9.0932149887085
2800 9.0351390838623
2900 8.96924066543579
3000 8.88618946075439
3100 8.86055088043213
3200 8.8024640083313
3300 8.72353172302246
3400 8.67293977737427
3500 8.61935472488403
3600 8.58778953552246
3700 8.54957294464111
3800 8.49683046340942
3900 8.4539680480957
4000 8.36084985733032
4100 8.32567358016968
4200 8.29765272140503
4300 8.25561809539795
4400 8.17453384399414
4500 8.1458535194397
4600 8.12973737716675
4700 8.09222650527954
4800 8.05335712432861
4900 8.02867746353149
5000 7.97845077514648
};
\addplot [semithick, red, dashed]
table {%
0 15.8379044532776
100 14.4138803482056
200 13.5033450126648
300 12.8437976837158
400 12.4406046867371
500 11.999623298645
600 11.574360370636
700 11.2705249786377
800 11.1022562980652
900 10.8538613319397
1000 10.5621380805969
1100 10.4464998245239
1200 10.2392468452454
1300 10.0873513221741
1400 9.87186813354492
1500 9.72915267944336
1600 9.65822076797485
1700 9.60069990158081
1800 9.46238040924072
1900 9.42171335220337
2000 9.31886148452759
2100 9.26916646957397
2200 9.19739151000977
2300 9.09852695465088
2400 9.05330228805542
2500 9.07623910903931
2600 8.92517185211182
2700 8.87928581237793
2800 8.80629253387451
2900 8.76802110671997
3000 8.7296576499939
3100 8.72359657287598
3200 8.64823865890503
3300 8.58444690704346
3400 8.6061110496521
3500 8.557053565979
3600 8.50903272628784
3700 8.46388006210327
3800 8.38205337524414
3900 8.40620899200439
4000 8.37509918212891
4100 8.33791542053223
4200 8.30606079101562
4300 8.30045938491821
4400 8.29828596115112
4500 8.24817657470703
4600 8.23934698104858
4700 8.22913074493408
4800 8.23248529434204
4900 8.19022083282471
5000 8.13958072662354
};

\addplot [semithick, green]
table {%
0 38.2023773193359
100 15.6015992164612
200 13.6732912063599
300 12.4754676818848
400 11.4447741508484
500 10.58260679245
600 9.94944429397583
700 9.51057624816895
800 9.12305116653442
900 8.84782886505127
1000 8.53074026107788
1100 8.23262119293213
1200 8.03249430656433
1300 7.87337613105774
1400 7.70135951042175
1500 7.57082271575928
1600 7.46085524559021
1700 7.32682371139526
1800 7.20460987091064
1900 7.11075162887573
2000 7.06459784507751
2100 6.99941539764404
2200 6.93596959114075
2300 6.89539813995361
2400 6.85997796058655
2500 6.80863785743713
2600 6.75676274299622
2700 6.73455715179443
2800 6.69008660316467
2900 6.65438723564148
3000 6.61851453781128
3100 6.59494352340698
3200 6.56795501708984
3300 6.5483021736145
3400 6.50633001327515
3500 6.48784303665161
3600 6.45515012741089
3700 6.422043800354
3800 6.39844346046448
3900 6.38447332382202
4000 6.36628460884094
4100 6.35332131385803
4200 6.33613348007202
4300 6.3196496963501
4400 6.29823660850525
4500 6.28215932846069
4600 6.26988101005554
4700 6.25633263587952
4800 6.24229264259338
4900 6.21937489509583
5000 6.19241285324097
};
\addplot [semithick, green, dashed]
table {%
0 37.4056053161621
100 15.7039523124695
200 13.3792028427124
300 12.1997151374817
400 11.2492027282715
500 10.5341606140137
600 10.1082859039307
700 9.78000593185425
800 9.55889415740967
900 9.123863697052
1000 8.92560434341431
1100 8.68587350845337
1200 8.44948673248291
1300 8.38651084899902
1400 8.18477630615234
1500 8.06571316719055
1600 7.97053194046021
1700 7.8756799697876
1800 7.79354619979858
1900 7.68289756774902
2000 7.5648238658905
2100 7.52376961708069
2200 7.47944951057434
2300 7.41581654548645
2400 7.38376522064209
2500 7.33275842666626
2600 7.30716490745544
2700 7.2732720375061
2800 7.22716689109802
2900 7.15291094779968
3000 7.13519191741943
3100 7.08665442466736
3200 7.05361342430115
3300 7.04846549034119
3400 7.04741501808167
3500 6.94636130332947
3600 6.97468662261963
3700 6.93889689445496
3800 6.90382218360901
3900 6.88654828071594
4000 6.84202122688293
4100 6.81373953819275
4200 6.78430819511414
4300 6.77668476104736
4400 6.75182223320007
4500 6.7568883895874
4600 6.74480867385864
4700 6.7050576210022
4800 6.70132112503052
4900 6.64463424682617
5000 6.61574745178223
};
\addplot [semithick, blue]
table {%
0 7.79783415794373
100 7.53067898750305
200 7.39190125465393
300 7.31955170631409
400 7.28342175483704
500 7.24195408821106
600 7.17333745956421
700 7.09025263786316
800 7.05174279212952
900 7.00437140464783
1000 6.98398852348328
1100 6.95531868934631
1200 6.93817901611328
1300 6.9265251159668
1400 6.91431760787964
1500 6.88616991043091
1600 6.8702220916748
1700 6.85294222831726
1800 6.8404974937439
1900 6.82781958580017
2000 6.81853127479553
2100 6.81505584716797
2200 6.80794167518616
2300 6.79217958450317
2400 6.78005504608154
2500 6.76558065414429
2600 6.75907301902771
2700 6.75453066825867
2800 6.74665880203247
2900 6.73877882957458
3000 6.73138356208801
3100 6.72126650810242
3200 6.70713949203491
3300 6.69809985160828
3400 6.68959951400757
3500 6.68565082550049
3600 6.68030118942261
3700 6.65434718132019
3800 6.64733695983887
3900 6.63980555534363
4000 6.62373161315918
4100 6.61364126205444
4200 6.59970283508301
4300 6.5848069190979
4400 6.57368612289429
4500 6.5651707649231
4600 6.5544810295105
4700 6.54977798461914
4800 6.53911972045898
4900 6.53493857383728
5000 6.52776074409485
};
\addplot [semithick, blue, dashed]
table {%
0 8.36278820037842
100 8.13541173934937
200 7.96603584289551
300 7.95825242996216
400 7.87746810913086
500 7.80725312232971
600 7.80536293983459
700 7.78959202766418
800 7.73987722396851
900 7.64695453643799
1000 7.61348223686218
1100 7.57540655136108
1200 7.51986956596375
1300 7.4987416267395
1400 7.44718170166016
1500 7.40093660354614
1600 7.35620951652527
1700 7.26267385482788
1800 7.24293231964111
1900 7.2077362537384
2000 7.21719360351562
2100 7.18254733085632
2200 7.12223172187805
2300 7.10976099967957
2400 7.11280941963196
2500 7.09070515632629
2600 7.08851051330566
2700 7.07339119911194
2800 7.033198595047
2900 6.99519824981689
3000 6.98209667205811
3100 6.96941161155701
3200 6.96866893768311
3300 6.96586084365845
3400 6.94114017486572
3500 6.96115016937256
3600 6.95769762992859
3700 6.9291524887085
3800 6.92798089981079
3900 6.90830254554749
4000 6.91392612457275
4100 6.87990546226501
4200 6.89035129547119
4300 6.89124274253845
4400 6.86529874801636
4500 6.83938884735107
4600 6.80929756164551
4700 6.79515290260315
4800 6.78452706336975
4900 6.79374408721924
5000 6.7561719417572
};

\nextgroupplot[
legend cell align={left},
legend style={
  draw = none,  
  fill = none,
  text opacity=1,
  at={(-0.1,-0.35)},
  anchor=north,
  legend columns=-1,
},
tick align=outside,
tick pos=left,
title={airfoil},
x grid style={darkgray176},
xlabel={Generation},
xmin=-250, xmax=5250,
xtick style={color=black},
y grid style={darkgray176},
ymin=3.3, ymax=6,
ytick style={color=black}
]
\addplot [semithick, red]
table {%
0 5.79739427566528
100 5.52734041213989
200 5.29409170150757
300 5.15090465545654
400 5.04999446868896
500 4.98857021331787
600 4.88431310653687
700 4.84160542488098
800 4.77932977676392
900 4.71701335906982
1000 4.68494558334351
1100 4.65711069107056
1200 4.60682058334351
1300 4.58856678009033
1400 4.56462287902832
1500 4.52275204658508
1600 4.49166464805603
1700 4.4689245223999
1800 4.44683265686035
1900 4.43077754974365
2000 4.41045784950256
2100 4.38639569282532
2200 4.36871838569641
2300 4.35009169578552
2400 4.3475546836853
2500 4.32519102096558
2600 4.30027937889099
2700 4.29105520248413
2800 4.2716498374939
2900 4.253258228302
3000 4.23813343048096
3100 4.23030757904053
3200 4.22067499160767
3300 4.19863748550415
3400 4.18035626411438
3500 4.15921139717102
3600 4.14409708976746
3700 4.12481284141541
3800 4.11726188659668
3900 4.10560727119446
4000 4.10041832923889
4100 4.09687948226929
4200 4.09000062942505
4300 4.07960557937622
4400 4.07552409172058
4500 4.07197332382202
4600 4.06521534919739
4700 4.05791163444519
4800 4.05066180229187
4900 4.04619336128235
5000 4.03980684280396
};
\addplot [semithick, red, dashed, forget plot]
table {%
0 5.71204280853271
100 5.59495854377747
200 5.38811588287354
300 5.21098685264587
400 5.11557078361511
500 5.05256223678589
600 4.98727107048035
700 4.95742440223694
800 4.9007396697998
900 4.84773397445679
1000 4.82678818702698
1100 4.7962851524353
1200 4.75499486923218
1300 4.73851132392883
1400 4.71149039268494
1500 4.69736051559448
1600 4.67765235900879
1700 4.66384291648865
1800 4.64155077934265
1900 4.64416670799255
2000 4.62608790397644
2100 4.61113357543945
2200 4.60483145713806
2300 4.58329701423645
2400 4.57649779319763
2500 4.52351450920105
2600 4.51248240470886
2700 4.50220394134521
2800 4.49150943756104
2900 4.50116181373596
3000 4.48266267776489
3100 4.48181223869324
3200 4.47522354125977
3300 4.44180274009705
3400 4.44237804412842
3500 4.43761730194092
3600 4.42894434928894
3700 4.39739465713501
3800 4.38688826560974
3900 4.36860132217407
4000 4.36969327926636
4100 4.35610127449036
4200 4.3451189994812
4300 4.33284401893616
4400 4.33194160461426
4500 4.31669354438782
4600 4.30702567100525
4700 4.30332732200623
4800 4.29788064956665
4900 4.29522800445557
5000 4.30109405517578
};

\addplot [semithick, green]
table {%
0 124.218254089355
100 83.7174301147461
200 42.8233757019043
300 10.5078182220459
400 6.21789145469666
500 5.09791445732117
600 4.64809274673462
700 4.44296312332153
800 4.31309819221497
900 4.23836755752563
1000 4.14382266998291
1100 4.06932783126831
1200 4.02431726455688
1300 3.98230051994324
1400 3.94740355014801
1500 3.91880464553833
1600 3.89474618434906
1700 3.85010766983032
1800 3.83013260364532
1900 3.81181728839874
2000 3.77905094623566
2100 3.765021443367
2200 3.74394583702087
2300 3.72263050079346
2400 3.71144342422485
2500 3.70237529277801
2600 3.68280375003815
2700 3.66947555541992
2800 3.65842473506927
2900 3.65117108821869
3000 3.64828991889954
3100 3.6401332616806
3200 3.63072264194489
3300 3.6181378364563
3400 3.61182618141174
3500 3.60107457637787
3600 3.59338438510895
3700 3.58616161346436
3800 3.57916021347046
3900 3.5740669965744
4000 3.56890225410461
4100 3.56018054485321
4200 3.55601525306702
4300 3.55056571960449
4400 3.54339861869812
4500 3.53771424293518
4600 3.53306949138641
4700 3.52447128295898
4800 3.51031768321991
4900 3.50641238689423
5000 3.49562847614288
};
\addplot [semithick, green, dashed, forget plot]
table {%
0 124.320320129395
100 83.9848823547363
200 43.1290721893311
300 10.1672921180725
400 6.12715530395508
500 5.19562959671021
600 4.69774007797241
700 4.54593133926392
800 4.47142624855042
900 4.4011116027832
1000 4.30632328987122
1100 4.25847411155701
1200 4.20480751991272
1300 4.17247653007507
1400 4.12795090675354
1500 4.11758136749268
1600 4.08216714859009
1700 4.06819200515747
1800 4.03690576553345
1900 4.01868104934692
2000 4.00357520580292
2100 3.99091970920563
2200 3.95396363735199
2300 3.93856942653656
2400 3.91288936138153
2500 3.90342020988464
2600 3.88340485095978
2700 3.87398493289948
2800 3.86883461475372
2900 3.86073744297028
3000 3.83867251873016
3100 3.83378040790558
3200 3.81579840183258
3300 3.80462431907654
3400 3.79611098766327
3500 3.787757396698
3600 3.77334880828857
3700 3.76354420185089
3800 3.76081967353821
3900 3.74291670322418
4000 3.74822664260864
4100 3.73863625526428
4200 3.73384761810303
4300 3.73004019260406
4400 3.72290849685669
4500 3.72241604328156
4600 3.70616257190704
4700 3.6933650970459
4800 3.68367314338684
4900 3.69106638431549
5000 3.67891299724579
};
\addplot [semithick, blue]
table {%
0 4.6088011264801
100 4.59907603263855
200 4.58142447471619
300 4.56808018684387
400 4.55374622344971
500 4.55081534385681
600 4.54712390899658
700 4.52940130233765
800 4.5194616317749
900 4.51669430732727
1000 4.50545859336853
1100 4.50147318840027
1200 4.49845361709595
1300 4.49292707443237
1400 4.48695683479309
1500 4.48255944252014
1600 4.48143744468689
1700 4.47846484184265
1800 4.47620177268982
1900 4.47613501548767
2000 4.47596120834351
2100 4.47199702262878
2200 4.4650993347168
2300 4.46324896812439
2400 4.45429468154907
2500 4.44944763183594
2600 4.44496655464172
2700 4.4319167137146
2800 4.42844462394714
2900 4.42597985267639
3000 4.42203569412231
3100 4.41965293884277
3200 4.41796612739563
3300 4.41701531410217
3400 4.41623687744141
3500 4.41456937789917
3600 4.41086292266846
3700 4.40257930755615
3800 4.40021109580994
3900 4.39986205101013
4000 4.39462089538574
4100 4.38855481147766
4200 4.38449740409851
4300 4.38104939460754
4400 4.37646675109863
4500 4.36853742599487
4600 4.36611914634705
4700 4.35670971870422
4800 4.35342311859131
4900 4.35142493247986
5000 4.33947515487671
};
\addplot [semithick, blue, dashed, forget plot]
table {%
0 4.79684352874756
100 4.76968121528625
200 4.76209616661072
300 4.76498627662659
400 4.73749041557312
500 4.73366951942444
600 4.73031902313232
700 4.72097969055176
800 4.71736168861389
900 4.71046686172485
1000 4.70530295372009
1100 4.69464373588562
1200 4.6908745765686
1300 4.68384027481079
1400 4.67671275138855
1500 4.67457008361816
1600 4.67366480827332
1700 4.67056274414062
1800 4.67250967025757
1900 4.6707022190094
2000 4.65066027641296
2100 4.64446902275085
2200 4.63978791236877
2300 4.63397836685181
2400 4.63295555114746
2500 4.62325882911682
2600 4.61249184608459
2700 4.60048580169678
2800 4.59782671928406
2900 4.60034418106079
3000 4.58914303779602
3100 4.5834972858429
3200 4.5827910900116
3300 4.57964181900024
3400 4.57136178016663
3500 4.5621132850647
3600 4.55661487579346
3700 4.55323314666748
3800 4.54517412185669
3900 4.53986716270447
4000 4.5377254486084
4100 4.53129601478577
4200 4.528639793396
4300 4.52312326431274
4400 4.52228283882141
4500 4.51748633384705
4600 4.52180409431458
4700 4.51557350158691
4800 4.51302146911621
4900 4.49875378608704
5000 4.49277758598328
};
\end{groupplot}
\end{tikzpicture}

\definecolor{green}{RGB}{0,128,0}

\addlegendimageintext{draw=red, line width=1.5pt} No AprT
\addlegendimageintext{draw=green, line width=1.5pt} Half AprT
\addlegendimageintext{draw=blue, line width=1.5pt} Full AprT

\hspace{5mm}
\addlegendimageintext{draw=black, line width=1.5pt} Train
\addlegendimageintext{draw=black, line width=1.5pt, densely dashed} Test

    \caption{Comparison of the impact of AprT on the evolutionary process, showing RMSE performance on the training and test sets across all datasets.}
    \label{fig: pre train}
\end{figure}

%% file: figures/post_train_bp.tex
\begin{figure*}[!h]
    \centering
    \begin{tikzpicture}
        \begin{groupplot}[
            width=38mm,
            height=38mm,
            group style={
                group size=4 by 1,
                horizontal sep=7mm,
                vertical sep=7mm,
                xlabels at=edge bottom,
                ylabels at=edge left
            },
            gridded,
            noinnerticks,
        ]

        \nextgroupplot[boxplot,
                boxplot/draw direction=y,
                xticklabels={},
                ylabel = RMSE,
                title={bioav},
                ]
        \boxplot[bpcolor=col11]{figures/boxplots/bioav.txt}{NEVO-GSPT};
        \boxplot[bpcolor=col12]{figures/boxplots/bioav.txt}{ApT};

        \nextgroupplot[boxplot,
                boxplot/draw direction=y,
                ymin=1000, ymax=5500,
                xticklabels={},
                title={ld50},
                ]
        \boxplot[bpcolor=col11]{figures/boxplots/ld50.txt}{NEVO-GSPT};
        \boxplot[bpcolor=col12]{figures/boxplots/ld50.txt}{ApT};

        \nextgroupplot[boxplot,
                boxplot/draw direction=y,
                xticklabels={},
                title={concrete\_strength},
                ]
        \boxplot[bpcolor=col11]{figures/boxplots/concrete_strength.txt}{NEVO-GSPT};
        \boxplot[bpcolor=col12]{figures/boxplots/concrete_strength.txt}{ApT};

        \nextgroupplot[boxplot,
                boxplot/draw direction=y,
                xticklabels={},
                title={airfoil},
                ]
        \boxplot[bpcolor=col11]{figures/boxplots/airfoil.txt}{NEVO-GSPT};
        \boxplot[bpcolor=col12]{figures/boxplots/airfoil.txt}{ApT};

        \end{groupplot}    
    \end{tikzpicture} 
    
    \addlegendimageintext{draw=col11, fill=col11!10, area legend, line width=1pt} NEVO-GSPT
    \addlegendimageintext{draw=col12, fill=col12!10, area legend, line width=1pt} A posteriori Training
    
    \caption{Comparison of the impact of ApoT on the evolutionary process, showing
RMSE performance on the test set across all datasets.
   }
    \label{fig:bp apot}
\end{figure*}
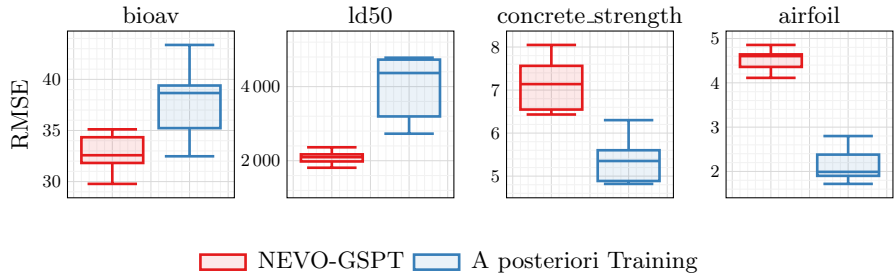

%% file: figures/final_results_bp.tex
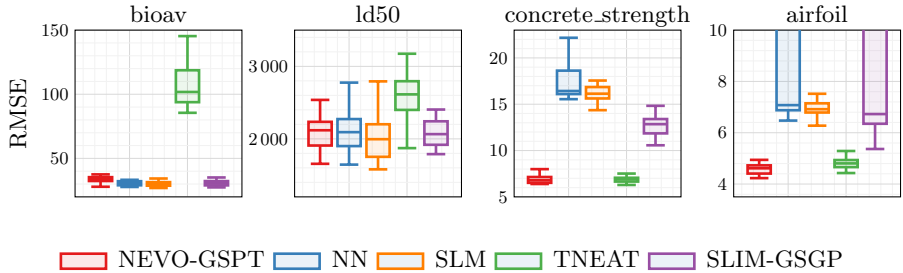
\begin{figure*}[!h]
    \centering
    \begin{tikzpicture}
        \begin{groupplot}[
            width=38mm,
            height=38mm,
            group style={
                group size=4 by 1,
                horizontal sep=7mm,
                vertical sep=7mm,
                xlabels at=edge bottom,
                ylabels at=edge left
            },
            gridded,
            noinnerticks,
        ]

        \nextgroupplot[boxplot,
                boxplot/draw direction=y,
                ymin=20, ymax=150,
                xticklabels={},
                ylabel = RMSE,
                title={bioav},
                ]
        \boxplot[bpcolor=col11]{figures/final_results/bioav.txt}{NEVO-GSPT};
        \boxplot[bpcolor=col12]{figures/final_results/bioav.txt}{NN};
        \boxplot[bpcolor=col15]{figures/final_results/bioav.txt}{SLM};
        \boxplot[bpcolor=col13]
        {figures/final_results/bioav.txt}{TNEAT};
        \boxplot[bpcolor=col14]{figures/final_results/bioav.txt}{SLIM};
        
        \nextgroupplot[boxplot,
                boxplot/draw direction=y,
                ymin=1200, ymax=3500,
                xticklabels={},
                title={ld50},
                ]
        \boxplot[bpcolor=col11]{figures/final_results/ld50.txt}{NEVO-GSPT};
        \boxplot[bpcolor=col12]{figures/final_results/ld50.txt}{NN};
        \boxplot[bpcolor=col15]{figures/final_results/ld50.txt}{SLM};
        \boxplot[bpcolor=col13]
        {figures/final_results/ld50.txt}{TNEAT};
        \boxplot[bpcolor=col14]{figures/final_results/ld50.txt}{SLIM};

        \nextgroupplot[boxplot,
                boxplot/draw direction=y,
                ymin=5, ymax=23,
                xticklabels={},
                title={concrete\_strength},
                ]
        \boxplot[bpcolor=col11]{figures/final_results/concrete_strength.txt}{NEVO-GSPT};
        \boxplot[bpcolor=col12]{figures/final_results/concrete_strength.txt}{NN};
        \boxplot[bpcolor=col15]{figures/final_results/concrete_strength.txt}{SLM};
        \boxplot[bpcolor=col13]
        {figures/final_results/concrete_strength.txt}{TNEAT};
        \boxplot[bpcolor=col14]{figures/final_results/concrete_strength.txt}{SLIM};

        \nextgroupplot[boxplot,
                boxplot/draw direction=y,
                ymin=3.5, ymax=10,
                xticklabels={},
                title={airfoil},
                ]
        \boxplot[bpcolor=col11]{figures/final_results/airfoil.txt}{NEVO-GSPT};
        \boxplot[bpcolor=col12]{figures/final_results/airfoil.txt}{NN};
        \boxplot[bpcolor=col15]{figures/final_results/airfoil.txt}{SLM};
        \boxplot[bpcolor=col13]
        {figures/final_results/airfoil.txt}{TNEAT};
        \boxplot[bpcolor=col14]{figures/final_results/airfoil.txt}{SLIM};

        \end{groupplot}    
    \end{tikzpicture} 
    
    \addlegendimageintext{draw=col11, fill=col11!10, area legend, line width=1pt} NEVO-GSPT
    \addlegendimageintext{draw=col12, fill=col12!10, area legend, line width=1pt} NN
    \addlegendimageintext{draw=col15, fill=col12!10, area legend, line width=1pt} SLM
    \addlegendimageintext{draw=col13, fill=col12!10, area legend, line width=1pt} TNEAT
    \addlegendimageintext{draw=col14, fill=col12!10, area legend, line width=1pt} SLIM-GSGP
    
    \caption{Comparison of the test performance of \gls{ngspt} against a set of baslines, presented as RMSE.
   }
    \label{fig:bp}
\end{figure*}

%% file: figures/size.tex
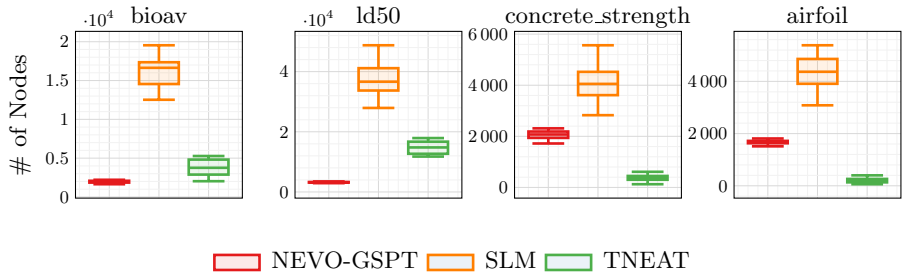
\begin{figure*}[!h]
    \centering
    \begin{tikzpicture}
        \begin{groupplot}[
            width=38mm,
            height=38mm,
            group style={
                group size=4 by 1,
                horizontal sep=7mm,
                vertical sep=7mm,
                xlabels at=edge bottom,
                ylabels at=edge left
            },
            gridded,
            noinnerticks,
        ]

        \nextgroupplot[boxplot,
                boxplot/draw direction=y,
                xticklabels={},
                ylabel = \# of Nodes,
                title={bioav},
                ]
        \boxplot[bpcolor=col11]{figures/sizes/bioav.txt}{NEVO-GSPT};
        \boxplot[bpcolor=col15]{figures/sizes/bioav.txt}{SLM};
        \boxplot[bpcolor=col13]
        {figures/sizes/bioav.txt}{TNEAT};
        
        \nextgroupplot[boxplot,
                boxplot/draw direction=y,
                xticklabels={},
                title={ld50},
                ]
        \boxplot[bpcolor=col11]{figures/sizes/ld50.txt}{NEVO-GSPT};
        \boxplot[bpcolor=col15]{figures/sizes/ld50.txt}{SLM};
        \boxplot[bpcolor=col13]
        {figures/sizes/ld50.txt}{TNEAT};

        \nextgroupplot[boxplot,
                boxplot/draw direction=y,
                xticklabels={},
                title={concrete\_strength},
                ]
        \boxplot[bpcolor=col11]{figures/sizes/concrete_strength.txt}{NEVO-GSPT};
        \boxplot[bpcolor=col15]{figures/sizes/concrete_strength.txt}{SLM};
        \boxplot[bpcolor=col13]
        {figures/sizes/concrete_strength.txt}{TNEAT};

        \nextgroupplot[boxplot,
                boxplot/draw direction=y,
                xticklabels={},
                title={airfoil},
                ]
        \boxplot[bpcolor=col11]{figures/sizes/airfoil.txt}{NEVO-GSPT};
        \boxplot[bpcolor=col15]{figures/sizes/airfoil.txt}{SLM};
        \boxplot[bpcolor=col13]
        {figures/sizes/airfoil.txt}{TNEAT};

        \end{groupplot}    
    \end{tikzpicture} 
    
    \addlegendimageintext{draw=col11, fill=col11!10, area legend, line width=1pt} NEVO-GSPT
    \addlegendimageintext{draw=col15, fill=col12!10, area legend, line width=1pt} SLM
    \addlegendimageintext{draw=col13, fill=col12!10, area legend, line width=1pt} TNEAT
    
    \caption{Comparison of the number of nodes of the individuals evolved by \gls{ngspt} against a set of baselines.
   }
    \label{fig:bp size}
\end{figure*}